\documentclass[fleqn,10pt]{wlscirep}
\usepackage[utf8]{inputenc}
\usepackage[T1]{fontenc}
\usepackage{soul, color}
\usepackage{multirow}
\usepackage[ruled,vlined]{algorithm2e}
\usepackage{colortbl}
\usepackage{tabu}
\usepackage{bm}

\definecolor{green}{rgb}{0.9,0.9,0.9}

\SetKwInput{kwLoss}{Loss}
\SetKwInput{kwInit}{Init Neurons' states}

\title{Accurate and efficient time-domain classification with adaptive spiking recurrent neural networks}

\author[1,*]{Bojian Yin}
\author[2]{Federico Corradi}
\author[1,3,4]{Sander M. Boht\'e}
\affil[1]{CWI, Machine Learning group, Amsterdam, The Netherlands}
\affil[2]{Stichting IMEC Netherlands, Holst Centre, Eindhoven, The Netherlands}
\affil[3]{Univ of Amsterdam, Faculty of Science, Amsterdam, The Netherlands}
\affil[4]{Rijksuniversiteit Groningen, Faculty of Science and Engineering, Groningen, The Netherlands}
\affil[*]{byin@cwi.nl (corresponding author)}

\begin{abstract}
Inspired by more detailed modeling of biological neurons, Spiking neural networks (SNNs) have been investigated both as more biologically plausible  and potentially more powerful models of neural computation, and also with the aim of extracting biological neurons' energy efficiency; the performance of such networks however has remained lacking compared to classical artificial neural networks (ANNs). Here, we demonstrate how a novel surrogate gradient combined with recurrent networks of tunable and adaptive spiking neurons yields state-of-the-art for SNNs on challenging benchmarks in the time-domain, like speech and gesture recognition. This also exceeds the performance of standard classical recurrent neural networks (RNNs) and approaches that of the best modern ANNs. As these SNNs exhibit sparse spiking, we show that they theoretically are one to three orders of magnitude more computationally efficient compared to RNNs with comparable performance. Together, this positions SNNs as an attractive solution for AI hardware implementations. 
\end{abstract}
\begin{document}

\flushbottom
\maketitle


\thispagestyle{empty}



\section*{Introduction}

\begin{figure*}[hbt!]
\centering
\includegraphics[width=\textwidth]{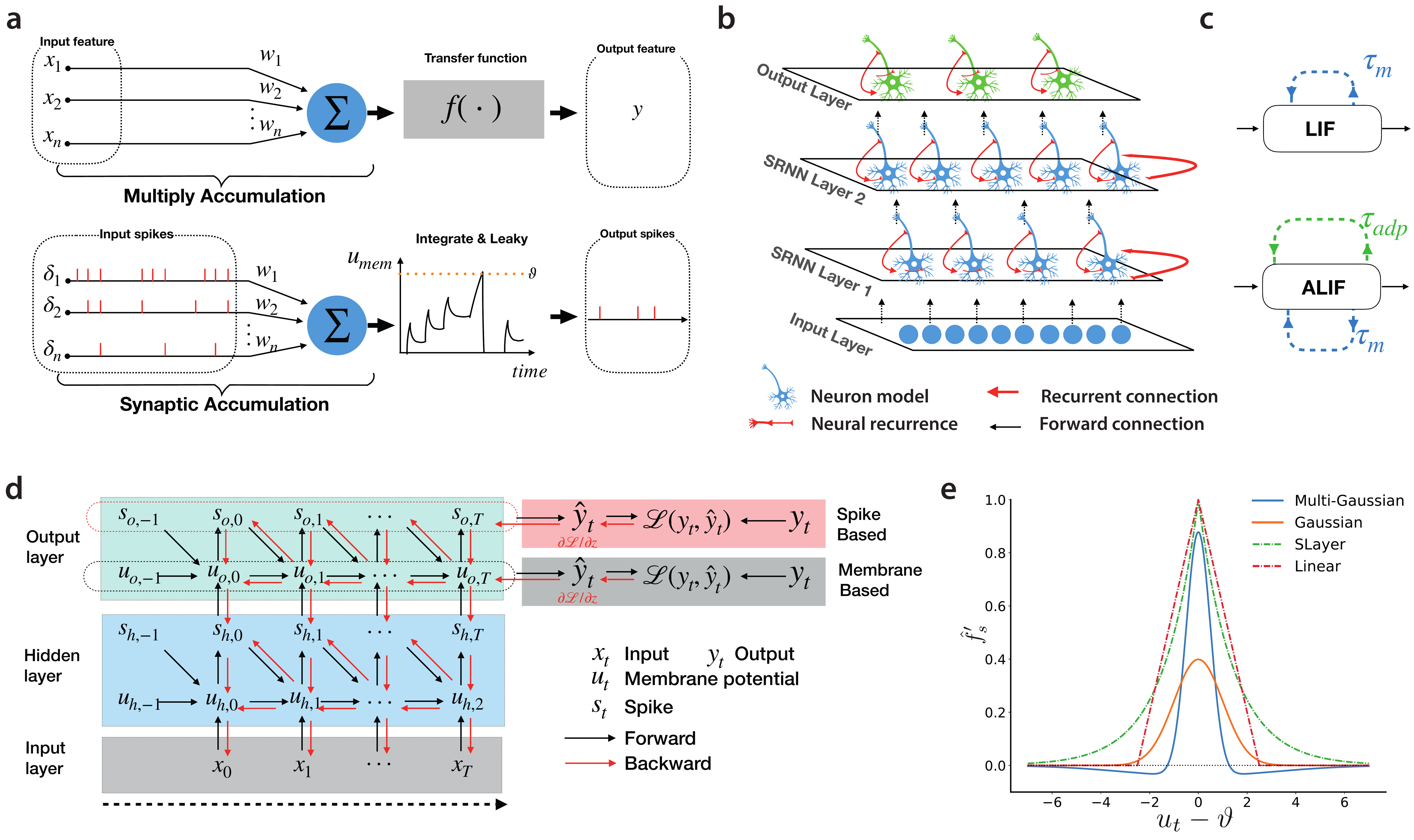}
\caption{\textbf{a}, Top: a classical artificial neural unit computes a weighted sum over input activations and then computes an output activation from this sum using a non-linear transfer function $f()$. Time is modeled as iterated recomputation of the network graph. Bottom: Spiking neurons receive spikes that are weighted and added to the internal state (membrane potential) that further develops through time following differential equations. When the membrane potential crosses a threshold, a spike is emitted and the potential is reset. \textbf{b}, Example architecture of a Spiking Recurrent Neural Network: an input layer projects to a layer of recurrently connected spiking neurons.  Recurrent layers then project to a read-out layer. Multiple recurrent layers can be connected in a feedforward fashion, here shown for two recurrent layers. \textbf{c}, The decaying threshold and membrane potential of the LIF and ALIF neurons can be modeled as an internal state induced by self-recurrency. \textbf{d}, Roll-out of the computational graph of a spiking neuron as used for backpropagation-through-time. \textbf{e}, Illustration of different surrogate gradient functions $\hat{f}'_s$ as a function of the neuron's membrane potential and threshold.
}
\label{fig:panel1}
\end{figure*}

The success of brain-inspired deep learning in AI is naturally focusing attention back onto those inspirations and abstractions from neuroscience \cite{hassabis2017neuroscience}. One such example is the abstraction of the sparse, pulsed and event-based nature of communication between biological neurons into neural units that communicate real values at every iteration or timestep of evaluation, taking the rate of firing of biological spiking neurons as an analog value (Figure \ref{fig:panel1}\textbf{a}). Spiking neurons, as more detailed neural abstractions, are theoretically more powerful compared to analog neural units \cite{maass1997networks} as they allow the relative timing of individual spikes to carry significant information. A real-world example in nature is the efficient sound localization in animals like Barn Owl's using precise spike-timing \cite{gerstner1996neuronal}. The sparse and binary nature of communication similarly has the potential to drastically reduce energy consumption in specialized hardware, in the form of neuromorphic computing \cite{davies2018loihi}. 

Since their introduction, numerous approaches to learning in spiking neural networks have been developed \cite{bohte2000spikeprop,shrestha2018slayer,zenke2018superspike,kheradpisheh2018stdp,falez2019multi}. 
All such approaches define how input signals are transduced into sequences of spikes, and how output spike-trains are interpreted with respect to goals, learning rules, or loss functions.
For supervised learning, approaches that calculate the gradient of weights with respect to the loss have to deal with the discontinuous nature of the spiking mechanism inside neurons. Local linearized approximations like SpikeProp \cite{bohte2000spikeprop} can be generalized to approximate ``surrogate'' gradients \cite{neftci2019surrogate}, or even calculated exactly in special cases \cite{wunderlich2020eventprop}. 
The use of surrogate gradients in particular has recently resulted in rapidly improving performance on select benchmarks, closing the performance gap with conventional deep learning approaches for smaller image recognition tasks like CIFAR10 and (Fashion) MNIST, and demonstrating improved performance on temporal tasks like TIMIT speech recognition \cite{bellec2020solution}. Still, spiking neural networks (SNNs) have struggled to demonstrate a clear advantage compared to classical artificial neural networks (ANNs) \cite{Sengupta2019-tg,roy2019towards}. 


Here, we introduce Spiking Recurrent Neural Networks (SRNNs), networks that include recurrently connected layers of spiking neurons (Figure \ref{fig:panel1}\textbf{b}). We demonstrate how these networks can be trained to high performance on hard benchmarks, exceeding existing state-of-the-art in SNNs, and approaching or exceeding state-of-the-art in classical recurrent artificial neural networks. The high-performance in SRNNs is achieved by applying back-propagation-through-time (BPTT)\cite{werbos1990backpropagation} to spiking neurons using a novel Multi-Gaussian surrogate gradient and using adaptive spiking neurons where the internal time-constant parameters are co-trained with network weights. The Multi-Gaussian surrogate gradient is constructed to include negative slopes, inspired by the ELU activation function \cite{clevert2015fast}: we find that the Multi-Gaussian surrogate gradient consistently outperforms other existing surrogate gradients. Similarly, co-training the internal time-constants of adaptive spiking neurons proved consistently beneficial. We demonstrate that these ingredients jointly improve performance to a competitive level while maintaining sparse average network activity. 

We demonstrate the superior performance of SRNNs for well-known benchmarks that have an inherent temporal dimension, like ECG wave-pattern classification, speech (Google Speech Commands, TIMIT), radar gesture recognition (SoLi), and classical hard benchmarks like sequential MNIST and its permuted variant. We find that the SRNNs need very little communication, with the average spiking neuron emitting a spike once every 3 to 30 timesteps depending on the task. Calculating the theoretical energy cost of computation, we then show that in SRNNs cheap Accumulate (AC) operations dominate over more expensive Multiply-Accumulate (MAC) operations. Based on relative MAC vs. AC energy cost \cite{roy2019towards,Sengupta2019-tg}, we argue that these sparsely spiking SRNNs have an energy advantage ranging from one to three orders of magnitude over RNNs and ANNs with comparable accuracy, depending on network and task complexity. 




\subsection*{Spiking Recurrent Neural Networks}
We focus here on multi-layer networks of recurrently connected spiking neurons, as illustrated in Figure \ref{fig:panel1}\textbf{b}; variations include networks that receive bi-directional input (bi-SRNNs; Figure \ref{fig:compcost_SI}\textbf{a}).

{\bf Spiking neurons} are derived from models that capture the behavior of real biological neurons \cite{gerstner2002spiking}. While biophysical models like the Hodgkin-Huxley model are accurate, they are also costly to compute\cite{izhikevich2003simple}. Phenomenological models like the Leaky-integrate-and-fire ({\bf LIF}) neuron model trade-of levels of biological realism for interpretability and reduced computational cost: the LIF neuron model integrates input current in a leaky fashion and emits a spike when its membrane potential crosses its threshold from below, after which the membrane potential is reset to the reset membrane potential; the current leak is determined by a decay time-constant $\tau_m$. 

As an exceedingly simple spiking neuron model, the LIF neuron lacks much of the complex behavior of real neurons, including responses that exhibit longer history dependency like spike-rate adaptation \cite{izhikevich2003simple}. Bellec et al.~\cite{bellec2018long} demonstrated how using a spiking neuron model that uses a generic form of adaptation improved performance in their SNNs. In this adaptive LIF ({\bf ALIF}) neuron, the LIF neuron model is augmented with an adaptive threshold that is increased after each emitted spike, and which then decays exponentially with time-constant $\tau_{adp}$. Both LIF and ALIF neurons can be thought of as neural units with self-recurrency, as illustrated in Figure \ref{fig:panel1}\textbf{c}.


\paragraph{BPPT, Surrogate-Gradient and Multi-Gaussian}




Given a loss-function $\mathcal{L}(t | { \theta})$ defined over neural activity at a particular time $t$, the error-backpropagation-through-time (BPTT) algorithm \cite{werbos1990backpropagation} updates network parameters ${\theta}$ in the direction that minimizes the loss by computing the partial gradient $\partial \mathcal{L}(t)/ \partial { \theta}$ using the chain-rule. Here, the parameters ${ \theta}$ include both the synaptic weights and the respective neural time-constants. In recurrently connected networks, past neural activations influence the current loss and by unrolling the network the contribution of these past activations to a current loss is accounted for. The roll-out of network activity through which the gradient is computed is illustrated in Figure \ref{fig:panel1}\textbf{d}. 

\begin{figure}[ht!]
\centering
\includegraphics[width=.65\textwidth]{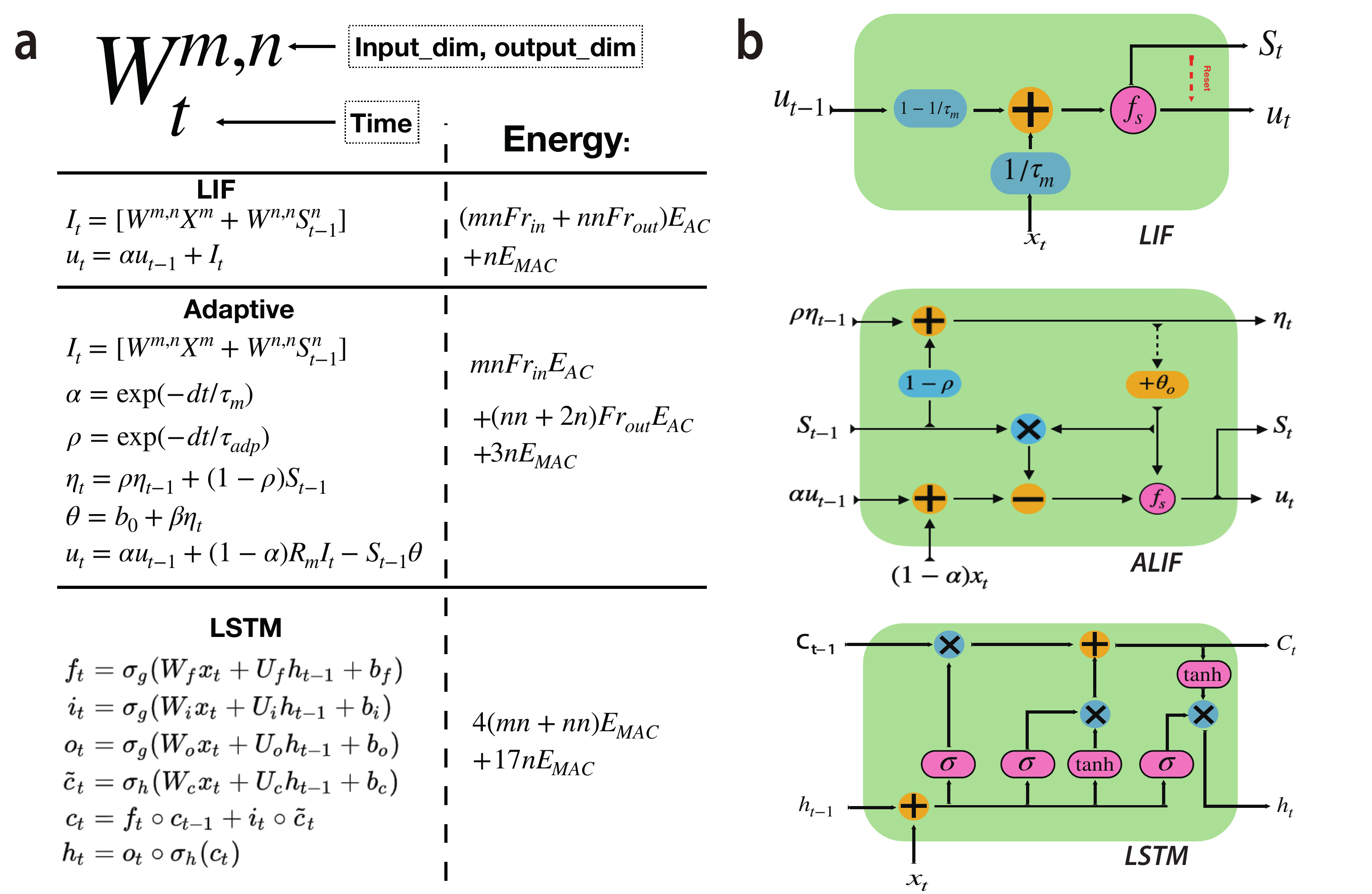}
\caption{\textbf{a}, Theoretical energy computation of different layers. The computational complexity calculation follows \cite{hunger2005floating}. \textbf{b}, LIF, ALIF and LSTM internal operation schematic.}
\label{fig:compcost}
\end{figure}

The discontinuous nature of the spiking mechanism in spiking neurons makes it difficult to apply the chain-rule connecting the backpropagating gradient between neural output and neural input \cite{bohte2000spikeprop}; in practice, replacing the discontinuous gradient with a smooth gradient function, a ``{\bf surrogate gradient}'' has proven effective  \cite{bohte2011error,neftci2019surrogate,bellec2020solution} and has the added benefit of allowing the mapping of spiking neural networks to recurrent neural networks in optimized Deep Learning frameworks like PyTorch and Tensorflow \cite{neftci2019surrogate}. Multiple surrogate gradient functions have been proposed and evaluated, including Gaussian, linear\cite{bellec2018long} and SLayer \cite{shrestha2018slayer} functions; for these functions however, no significant differences in performance are reported\cite{neftci2019surrogate}. 

Inspired by the ELU\cite{clevert2015fast} and LeakyRelu\cite{} activation function, we here define the Multi-Guassian (MG) a novel surrogate gradient $\hat{f_s}'(\cdot)$ comprised of a weighted sum of multiple Guassians $\mathcal{N}$ where the hyperparameter $h$ and $s$ are chosen such that the Multi-Guassian contains negative parts:
\begin{align}
\hat{f_s}'(u_{t}|\vartheta) = &  (1+h)\mathcal{N}(u_t|\vartheta,\sigma^2) - h\mathcal{N}(u_t|\sigma,(s \sigma)^2) -  h\mathcal{N}(u_t|-\sigma,(s \sigma)^2),    
\end{align}
where $u_t$ is the spiking neuron's membrane potential and $\vartheta$ its internal threshold. 
The negative parts of the ELU and Leaky-RELU functions are thought to alleviate the ``dying ReLU'' problem \cite{lu2019dying} where the sum of a neuron's input is negative for all inputs, and the neuron effectively does not participate in the network computation. The shape of the Multi-Gaussian (MG) and various other surrogate gradient functions is illustrated in Figure \ref{fig:panel1}\textbf{e}.

\paragraph{Computational Cost} 
To estimate the efficiency of SNNs and compare them to ANNs, we calculate the number of computations required in terms of accumulation (AC) and multiply-and-accumulate (MAC) operations\cite{wong2020tinyspeech}. We do this for an SRNN network with LIF or ALIF neurons and compare to a complex recurrent ANN structure like an LSTM\cite{hochreiter1997long} in Figure \ref{fig:compcost} -- for other ANNs, see Figure \ref{fig:compcost_SI}b.
In ANNs, the contribution from one neuron to another requires a MAC for every timestep, multiplying each input activation with the respective weight before adding to the internal sum. In contrast, for a spiking neuron a transmitted spike requires only an AC at the target neuron, adding the weight to the potential, and where spike inputs may be quite sparse. In addition, the spiking neuron's internal state requires updating every timestep at the cost of several MACs depending on the spiking neuron model complexity \cite{roy2019towards}. As calculating MACs is much more energetically expensive compared to ACs (e.g., 31x on 45nm CMOS\cite{horowitz20141}), the relative efficiency of SNNs is determined by the number of connections times activity sparsity and the spiking neuron model complexity.  Additionally, we remark that in hardware, multiplication circuits require substantially more die area compared to addition circuits \cite{ludgate1982proposed}.  

\section*{Results}

\begin{table}[]
\centering
\caption{\textbf{Comparison of SRNN performance to respective RNN and SNN state-of-the art accuracy (Acc.).}}
\begin{tabular}{|l|l|l|l|l|l|l|l|}
\hline
Task                     & Network                                & Method                                          & Acc.                                  & Task                   & Network                                                       & Method                                                    & Acc.                                                     \\ \hline
\multirow{2}{*}{ECG}     & \em RNN-SoTa            & Bi-LSTM                                         & 80.8\%                                   & \multirow{5}{*}{SSC}   & \em  RNN-SoTa                                  & LSTM\cite{cramer2019heidelberg}          & 73.1\%                                                       \\ \cline{2-4} \cline{6-8} 
                         & \cellcolor{green}\bf SRNN & \cellcolor{green}\bf Ours          & \cellcolor{green}\bf 85.9\%  &                        & \em CNN-SoTa                                   & CNN\cite{cramer2019heidelberg}           & 77.7\%                                                       \\ \cline{1-4} \cline{6-8} 
\multirow{4}{*}{SMNIST}  & \em RNN-SoTa            & IndRNN\cite{li2018independently}        & 99.5\%                                   &                        & SNN-base                                                      & LIF \cite{cramer2019heidelberg}          & 50.1\%                                                      \\ \cline{2-4} \cline{6-8} 
                         & \em RNN                 & LSTM\cite{arjovsky2016unitary} & 98.2\%                                    &                        & SRNN-SoTA                        &      SNN\cite{perez2021neural}               & 60.1\%                      \\ \cline{2-4} \cline{6-8} 
                         & SRNN-SoTa                              & LSNN\cite{bellec2018long}      & 96.4\%                                    &                        &\cellcolor{green}\bf SRNN                        & \cellcolor{green}\bf Ours                    & \cellcolor{green}\bf 74.2\%                                                              \\ \cline{2-8} 
                         & \cellcolor{green}\bf SRNN & \cellcolor{green}\bf Ours          & \cellcolor{green}\bf 98.7\% & \multirow{3}{*}{SoLi}  & \em CNN-SoTa                                   & CNN\cite{wang2016interacting}            & 77.7\%                                                      \\ \cline{1-4} \cline{6-8} 
\multirow{3}{*}{PSMNIST} & \em  RNN-SoTa           & IndRNN\cite{li2018independently}        & 97.2\%                                    &                        & RNN-SoTa                                                      & CNN+LSTM\cite{wang2016interacting}       & 87.2\%                                                      \\ \cline{2-4} \cline{6-8} 
                         & \em  RNN                & LSTM\cite{arjovsky2016unitary} & 88\%                                      &                        & \cellcolor{green}\bf SRNN                        & \cellcolor{green}\bf Ours                    & \cellcolor{green}\bf 91.9\%                     \\ \cline{2-8} 
                         & \cellcolor{green}\bf SRNN & \cellcolor{green}\bf Ours          & \cellcolor{green}\bf 94.3\% & \multirow{4}{*}{GSC}   & \em  RNN-SoTa                                  & Att RNN\cite{de2018neural}               & 95.6\%                                                       \\ \cline{1-4} \cline{6-8} 
\multirow{6}{*}{SHD}     & \em  RNN-SoTa           & Bi-LSTM                                         & 87.2\%                                   &                        & \em CNN-SoTa                                                      & TinySpeech\cite{wong2020tinyspeech}            & 92.4\%                                                 \\ \cline{2-4} \cline{6-8} 
                         & \em  CNN-SoTa           & CNN\cite{cramer2019heidelberg} & 92.4\%                                    &                        & SNN-SoTa                                                      & LSNN\cite{bellec2020solution}            & 91.2  \%                   \\ \cline{2-4} \cline{6-8} 
                         & SNN-base                               & LIF\cite{cramer2019heidelberg} & 71.4\%                                    &                        &\cellcolor{green}\bf SRNN                        & \cellcolor{green}\bf Ours                    & \cellcolor{green}\bf 92.1\%                                                             \\ \cline{2-8} 
                         &SNN                               & SNN\cite{zenke2020remarkable} & 82.2\%                                       & \multirow{3}{*}{TIMIT} & \em  RNN-SoTa                                  & Bi-LSTM\cite{graves2005framewise}        & 68.9\%                                                       \\ \cline{2-4} \cline{6-8} 
                         &SNN-SoTa                               & SNN\cite{perez2021neural} & 82.7\%                                       &                        & SNN-SoTa                                                      & LSNN\cite{bellec2018long}                & 65.4\%                                                       \\ \cline{2-4} \cline{6-8} 
                         & \cellcolor{green}\bf SRNN & \cellcolor{green}\bf Ours          & \cellcolor{green}\bf 90.4\%  &                        & \cellcolor{green}\bf Bi-SRNN & \cellcolor{green}\bf Ours & \cellcolor{green}\bf 66.1\% \\ \hline
\end{tabular}
\label{tab:sota_acc}
\end{table}

\paragraph{Tasks} Recurrent neural networks (RNNs) provide state-of-art performance in various sequential tasks that require memory \cite{shewalkar2019performance} typically in small and compact networks, and can operate in an online fashion. We distinguish two kinds of sequential tasks: streaming tasks, where many inputs map to many specified outputs (many-to-many), and classification tasks where an input sequence maps to a single output value (many-to-one). Sequential classification tasks can additionally be computed in an online fashion, where a classification is determined for each timestep.



\begin{figure}[ht!]
\centering

\includegraphics[width=\textwidth]{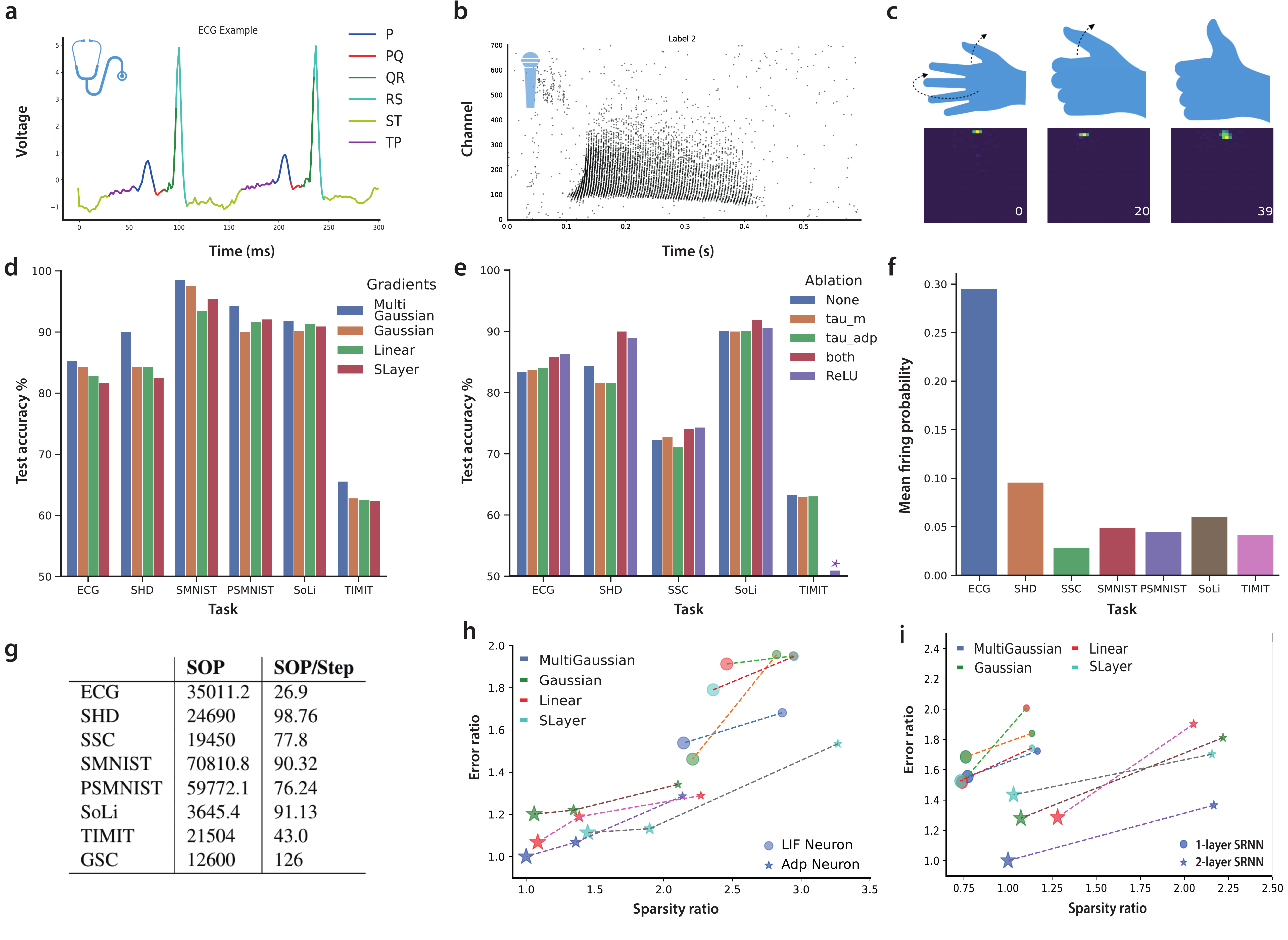}
\caption{Examples of \textbf{a}, a single ECG signal channel labeled for each timestep; \textbf{b}, the input spike-trains for the spoken number ``seven'' in the SHD dataset and \textbf{c}, example of gesture data -- the temporal evolution of the gesture (upper row) and the corresponding RDI (bottom row); \textbf{d}, Effects of various surrogate gradients on performance. \textbf{e}, Effects of training the time constant hyperparameters $\tau_m$ and $\tau_{adp}$; the legend denotes which hyperparameters are trained; ReLU denotes the non-spiking analog SRNN. \textbf{f}, The per timestep spike probability of the SRNNs on various tasks.  \textbf{g}, Total average Spike Operations (SOP) per sample and SOPs per sample per step (timestep/frame); \textbf{h},Effect of neuron types in terms of test accuracy and sparsity with various gradients (shown for SoLi dataset); the size of the nodes indicates the network size and the color of nodes represents the gradient type; \textbf{i}, Effect of the number of hidden recurrent layers on test accuracy and sparsity with various gradients (shown for SHD dataset).}
\label{fig:panel-2}
\end{figure}

We selected benchmark tasks that have an inherent temporal dimension can also be computed with neural networks of modest size to fit the dynamics and constraints of spiking neural networks. For these tasks, we trained several different SRNN network architectures with various gradients, hyperparameters, and spiking neuron models and compared them to classical and state-of-the-art RNN architectures.

The electrocardiogram ({\bf ECG}) \cite{laguna1997database} signal is composed of six different characteristic waveforms --~P, PQ, QR, RS, ST, and TP~-- whose shape and duration inform clinicians on the functioning of the cardiovascular system. The task requires the continuous recognition of all six waveforms, where we use signals from the QTDB dataset \cite{laguna1997database}. The ECG-wave labeling is an online and streaming task using only past information.

The sequential- and permuted-sequential {\bf S/PS-MNIST} datasets are standard sequence classification tasks of length 784  derived from the classical MNIST digit recognition task by presenting pixels one-at-time. The permuted version also first permutes each digit-class removing spatial information. 

The Spiking Heidelberg Dataset ({\bf SHD}) and Spiking Speech Command ({\bf SSC}) Dataset \cite{cramer2019heidelberg} are SNN specific sequence classification benchmarks comprised of audio converted into spike trains based on a detailed ear model. 

The {\bf SoLi} dataset\cite{wang2016interacting} gesture recognition task is comprised of a set of gestures in form of a sequence of radar returns collected the SoLi SoLid-state millimeter-wave radar sensor. We treat the SoLi task as both an online streaming and classification task by processing frames sequentially - we thus obtain two measures for the SoLi task, per-frame accuracy and whole sequence accuracy for streaming and classification respectively. 

Both the Google Speech Commands ({\bf GSC}) dataset\cite{warden2018speech} and the {\bf TIMIT} dataset \cite{garofolo1993timit} are classical speech recognition benchmarks where for TIMIT, we compute the Frame Error Rate (FER) and where, similar to \cite{bellec2020solution}, we apply a bidirectional architecture such that also future information is used to classify each frame (illustrated in Figure \ref{fig:compcost_SI}\textbf{a}). Samples from the ECG, SHD and SoLi datasets are shown in Figures \ref{fig:panel-2}\textbf{a-c}.


As shown in Table \ref{tab:sota_acc}, we find that these SRNNs achieve novel state-of-the-art for Spiking Neural Networks on all tasks, exceed conventional RNNs like LSTM models, and approach or exceeds the state-of-the-art of modern RNNs. We see that SRNNs substantially close the accuracy gap (SHD, SSC, GSC) compared to non-recurrent architectures like convolutional neural networks (CNNs) and Attention-based networks -- the latter networks however are typically comprised of many more neurons or parameters and cannot be computed in an online or streaming fashion. 

We plot the accuracy for the various tasks using different surrogate gradients in Figure \ref{fig:panel-2}\textbf{d}: while we see that there is little difference between previously developed gradients like Gaussian, Linear, and SLayer, we find that the Multi-Gaussian function consistently outperforms these gradients. We also find that independently of the surrogate gradient used, training the time-constants in the Adaptive LIF neurons consistently improves performance, as shown in the ablation study in Figure \ref{fig:panel-2}e: not training either $\tau_m$ or $\tau_{adp}$, or neither, reduces performance. Much of the power of the SRNNs seem to derive from their multi-layer recurrent and self-recurrent architecture. When we make the spiking neurons non-spiking by eliminating the spiking mechanism and communicating the RELU value of the membrane potential, we find that for almost all tasks we achieve performance that slightly exceeds that of the spiking SRNNs. 

The trained SRNNs communicate sparingly: most networks exhibit sparseness less than 0.1, and only the ECG task requires more spikes as it was tuned to use the smallest SRNN network (46 neurons). Sparseness of neural activity, expressed as average firing probability per timestep per neuron, is plotted in Figure \ref{fig:panel-2}\textbf{f}.
In general, we find that increasing network sizes improves accuracy while decreasing average sparsity (Figure \ref{fig:panel-2}\textbf{h-i}) -- though the total number of spikes used in the network increases. The total average number of spikes required per sample (SOPs) and per sample per step (SOP/step) for the highest performing SRNNs are given in Figure \ref{fig:panel-2}\textbf{g}. 

Plotting the performance of networks using either ALIF or LIF neurons, we find that using ALIF neurons consistently improves both performance and activity sparseness in the networks (Figure \ref{fig:panel-2}\textbf{h}). Similarly, splitting a single large recurrent layer into two layers of recurrently connected layers in the SRNN architecture improves both performance and sparsity in the SHD task (Figure \ref{fig:panel-2}\textbf{i}); we observed similar improvements in the other tasks. 

\begin{figure}[ht!]
\centering
\includegraphics[width=\textwidth]{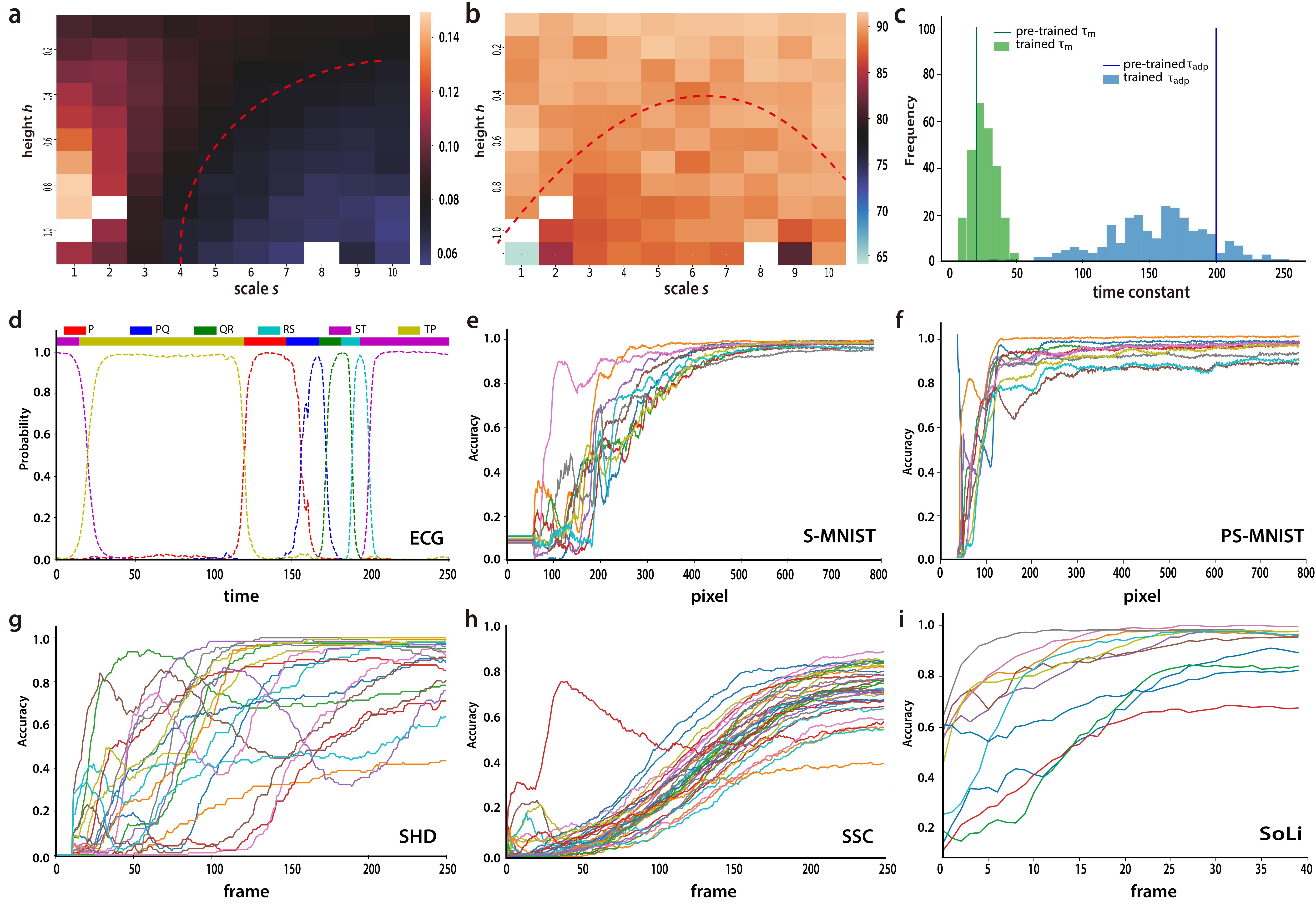}
\caption{\textbf{a,b}, Grid search for $h$ and $s$ parameters for the Multi-Gaussian surrogate gradient on SoLi dataset. The red dotted line highlights solutions with high accuracy \textbf{a}, and high sparsity \textbf{b}. \textbf{c}, evolution of spiking neuron time constants evolving before and after training \textbf{d}, example of ECG streaming classification: the prediction probability of each output label is calculated from the normalized output neurons' membrane potential (dashed lines, bottom); top: the color-coded true labels. \textbf{e-i} Temporal evolution of classification accuracy for the S-MNIST recognition task \textbf{e}, the PS-MNIST task \textbf{f}, the SHD recognition task \textbf{g},, the SSC dataset \textbf{h},, and for the SoLi dataset \textbf{i}. }

\label{fig:panel3}
\end{figure}

We carried out a grid search on the SoLi and SHD datasets for the $h$ and $s$ hyperparameters to determine the optimal parameter values for the Multi-Gaussian surrogate gradient. We find that there is a range of values where we can obtain both competitive accuracy and high sparsity (orange dotted line in Figure \ref{fig:panel3}\textbf{a-b}) -- we used a similar parameter search for the other tasks. The training procedure also substantially `learns' the time-constants for the respective tasks: as shown in Figure \ref{fig:panel3}\textbf{c} for the SHD task, starting from a tight distribution of time-constants, the spiking neurons in the trained network converge to using a wide variety of time-constants - the same effect is observed in the other tasks (not shown). 

The streaming and online nature of several of the tasks allows the network to make any-time decisions. Figure \ref{fig:panel3}\textbf{d} shows the classification for the various ECG waveforms for every timestep. When a new wave is presented, there is a brief delay before this class is correctly identified. In Figures \ref{fig:panel3}\textbf{e-i}, the average online classification performance is shown for the  S-MNIST, PS-MNIST, SHD, SSC, and SoLi datasets. We see that the S-MNIST and PS-MNIST digits can be recognized reliably quickly, while the SSC sounds require distinctly more time. The SHD sound recognition is much more erratic, and inspection of the data shows that this is caused by the various classes being placed at different times in the sound clip. Figure \ref{fig:panel3}\textbf{i} plots the accuracy as a function of the number of frames shown for the SoLi task. Most gestures can be recognized reliably already after having presented only 25 out of the 42 frames - comparing favorably with\cite{wang2016interacting}: the SRNN allows decisions to be made earlier and with better accuracy.

\begin{table}[]
\centering
\caption{\textbf{Comparison of SRNN energy consumption to respective RNN and SNN state-of-the-art accuracy.} Relative energy cost is calculated using the number of MACs and ACs required during inference with $Energy_{nn} = \sum_{t \in T} 3.1MAC+.1AC$ \cite{horowitz20141}; $fr$ denotes the average spiking probability in the SRNNs per timestep.
The Bidirectional-LSTM network $290^*$ contains 290 LSTM units. The accuracy in SoLi dataset is per frame accuracy. For GSC, the ReLu SRNN did not converge.
$^{**}$ For TIMIT, the complexity of comparably accurate networks was not available.}

\scalebox{0.8}{
\begin{tabular}{|l|l|l|l|l|l|l|l|l|l|l|l|l|l|}
\hline
\multirow{2}{*}{Task} & \multirow{2}{*}{Method} & \multirow{2}{*}{Network} & \multirow{2}{*}{Acc.} & \multicolumn{2}{l|}{Energy/Step} & \multirow{2}{*}{Ratio} & \multirow{2}{*}{Task} & \multirow{2}{*}{Method} & \multirow{2}{*}{Network} & \multirow{2}{*}{Acc} & \multicolumn{2}{l|}{Energy/Step} & \multirow{2}{*}{Ratio} \\ \cline{5-6} \cline{12-13}
                      &                         &                          &                      & MAC             & AC*fr          &                        &                       &                         &                          &                      & MAC             & AC*fr          &                        \\ \hline
\multirow{4}{*}{ECG}  & Bi-LSTM                 & $290^*$                     & 80.8                & 181.8K          &                & 1.7Kx                & \multirow{3}{*}{SHD}  & Bi-LSTM                 & 128+128+100              & 87.2                & 1.1M            &                & 1.7Kx                 \\ \cline{2-7} \cline{9-14} 
                      & ReLU                    & 4+36+6                   & 86.4                 & 1.9k            &                & 18x                     &                       & ReLu                    & 128+128                  & 88.9                & 142.6k          &                & 125x                 \\ \cline{2-7} \cline{9-14} 
                      & Ours (LIF)                     & 4+36+6                   & 49.7                 & 42              & .5k            & 0.5x                   &     & \cellcolor{green}Ours (ALIF)                    & \cellcolor{green}128+128                  & \cellcolor{green}87.8                & \cellcolor{green}788             & \cellcolor{green}10.7k          & \cellcolor{green}\bf 1x                     \\ \cline{2-14} 
                      & \cellcolor{green}Ours                    & \cellcolor{green}4+36+6                   & \cellcolor{green}85.9                 & \cellcolor{green}90              & \cellcolor{green}.5k            & \cellcolor{green}\bf 1x                     & \multirow{3}{*}{SoLi} & LSTM                    & 512+512                  & 77.7                & 2.7M            &                & 604x                 \\ \cline{1-7} \cline{9-14} 
S                     & ReLu                    & 64+256+256               & 99.0                & 157.3k          &                & 59x                  &                       & ReLu                    & 512+512                  & 79.6                & 1.1M            &                & 246x                   \\ \cline{2-7} \cline{9-14} 
MNIST                 & \cellcolor{green}Ours                    & \cellcolor{green}64+256+256               & \cellcolor{green}98.7                & \cellcolor{green}2k            & \cellcolor{green}20k           & \cellcolor{green}\bf 1x                     &                       & \cellcolor{green}Ours                    & \cellcolor{green}512+512                  & \cellcolor{green}79.8                & \cellcolor{green}3.1k            & \cellcolor{green}42.4k          & \cellcolor{green}1x                     \\ \hline
PS                    & ReLu                    & 64+256+256                & 93.5                & 157.3k          &                & 63x                  & \multirow{2}{*}{GSC}  & 
ReLu                    & 300+300                  &                      & 222.6k          &                & 167x  
\\ \cline{2-7} \cline{9-14} 
MNIST                 & \cellcolor{green}Ours                    & \cellcolor{green}64+256+256               & \cellcolor{green}94.3                & \cellcolor{green}2k            & \cellcolor{green}15.3k          & \cellcolor{green}\bf 1x                     &                       & \cellcolor{green}Ours                    & \cellcolor{green}300+300                  & \cellcolor{green}92.2                & \cellcolor{green}1k            & \cellcolor{green}10.1k          & \cellcolor{green}\bf 1x                  \\ \cline{1-14} 
\multirow{2}{*}{SSC}  & ReLu                    & 400+400                  & 74.4                & 766.6k          &                & 236x                 &                       
\multirow{2}{*}{TIMIT**}                & \multirow{2}{*}{Ours }                  & \multirow{2}{*}{256+61 }               & \multirow{2}{*}{66.1}                & \multirow{2}{*}{1.6k }           & \multirow{2}{*}{56.7k}          & \multirow{2}{*}{\bf 1x }                    \\ \cline{2-7}  

                      & \cellcolor{green}Ours                    & \cellcolor{green}400+400                  & \cellcolor{green}74.2                & \cellcolor{green}2.4k            & \cellcolor{green}26.1k          & \cellcolor{green}\bf 1x                     
                      &                &                   &                &               &        &          &                     \\ \hline
\end{tabular}
}
\label{tab:energy}
\end{table}

Given the relative AC and MAC energy cost from  \cite{horowitz20141,roy2019towards,kundu2021spike} and the computational complexity calculations from Figure \ref{fig:compcost}\textbf{a}, we plot in Table \ref{tab:energy} the relative energy efficiency of the various networks. We see that for the more complex tasks, SRNNs are theoretically at least 59x more energy efficient compared to RNNs at equivalent performance levels, where for most tasks the non-spiking (ReLU) SRNN compares most favourably. 
More classical RNN structures like LSTMs require many more parameters and operations, often being 1000x less efficient -- we also calculate similar estimates for other RNN structures in Table \ref{tab:sota-ratio}.

\section*{Discussion}


We showed how multi-layered recurrent network structures are able to achieve new state-of-the-art performance for SNNs on sequential and temporal tasks. This was accomplished by using adaptive spiking neurons with learned temporal dynamics trained with backpropagation-through-time using a novel surrogate gradient, the Multi-Gaussian, where the Multi-Gaussian gradient proved to consistently outperform the other  surrogate gradients. These results approach or equal the accuracy of conventional RNNs. When expressed in terms of computational operations, they demonstrate a decisive theoretical energy advantage of one to three orders of magnitude over conventional RNNs. This advantage furthermore increases for more complex tasks that required larger networks to solve accurately.

Neither the SRNNs nor the presented RNNs were optimized beyond accuracy and (for the SRNNs) sparsity: no optimizations like pruning and quantization were applied. When we compare the SRNN for the GSC task against the Attention-based CNN-network TinySpeech \cite{wong2020tinyspeech}, representing the recent state-of-the-art in efficiency-optimized speech recognition, we find that at an equivalent performance level, the SRNN still requires 19.6x fewer MACs, and where, unlike TinySpeech, the SRNN operates in an online and streaming fashion (data in Table \ref{tab:sota-ratio}).

We focused on temporal or sequential problems with relatively limited input dimensionality. With RNNs, such problems can be solved with relatively small neural networks and hold direct promise for implementation in ultra-low power EdgeAI solutions. This also was the reason for emphasizing streaming or online solutions where no or fixed preprocessing and buffering is required: problems where a temporal stream first has to be segmented and where these segments are then classified greatly increase the complexity of such solutions. As we demonstrated, most classification decisions could be made early with near-optimal accuracy.

Using surrogate-gradients, the BPTT-gradient in the SRNNs can be computed using standard deep learning frameworks, where we used PyTorch \cite{paszke2019pytorch}. With this approach, complicated architectures and spiking neuron models can be trained with state-of-the-art optimizers, regularizers, and visualization tools. At the same time, this approach is costly in terms of memory use and training time, as the computational graph is fully unrolled over all timesteps, and the abundant spatial and temporal sparsity is not exploited in the frameworks. This also limits the size of the networks to which this approach can be applied: for significantly larger networks, either dedicated hardware and/or sparsity optimized frameworks are needed\cite{zenkebohte2021}. Approximations to BPTT like eProp \cite{bellec2020solution} or alternative recurrent learning methods like RTRL\cite{zenke2021brain} may also help alleviate this limitation. 

We remark that the energy advantage of SRNNs we computed is theoretical: while the computational cost in terms of MACs is well-accepted \cite{kundu2021spike,wong2020tinyspeech}, this measure ignores real-world realities like the presence or absence of sufficient local memory, the cost of accessing memory, and the potential cost of routing spikes from one neuron to another. In many EdgeAI applications, the energy-cost of conventional sensors may also dominate the energy equation. At the same time, the numbers we present are unoptimized in the sense that other than optimizing the surrogate gradient for both sparsity and accuracy, we did not prune the networks or applied other standard optimization and quantization techniques. Substantial improvements here should be fairly straightforward. Training parameters of spiking neuron models in the SRNNs can be extended further to approaches that include parameterized short-term plasticity \cite{keijser2020interneuron} and more complicated spiking neuron models. 

The effectiveness of adjusting time-constant parameters to the task may also have implications for neuroscience: though effective time-constants of real spiking neurons are variable and dynamic \cite{gerstner2002spiking}, the benefit of training these parameters in SRNNs suggests these neural properties may be subject to learning processes in biology.

\section*{Methods}

In the SRNNs, the LIF spiking neuron is modeled as:
\begin{align}
\label{eq1:LIF}
& u_{t-1} = u_{t-1}(1-S_{t-1}) + u_r S_{t-1} \\
& u_t = u_{t-1}(1-1/\tau_m)+R_m I_t /\tau_m \\
& S_t = f_s(u_t,\vartheta) 
\end{align}
where $I_t = \sum_{t_i} w_i \delta(t_i) + I_{inj,t}$ is the input signal comprised of spikes at times $t_i$ weighted by weight $w_i$ and/or an injected current $I_{inj,t}$; $u$ is the neuron's membrane potential which decays exponentially with time-constant $\tau_m$, $\vartheta$ is the threshold, $R_m$ is the membrane resistance (which we absorb in the synaptic weights). The function $f_s(u_t,\vartheta)$ models the spike-generation mechanism as function of the threshold $\vartheta$, which is set to 1 when the neuron spikes and otherwise is 0 (where the approximating surrogate gradient is then $\hat{f}_s'(u_t,\vartheta)$). The value for the reset potential $u_r$ was set to zero. 
The ALIF neuron is similarly modeled as :
\begin{align}
\label{eq2:adapt}
&u_t = \alpha u_{t-1} + (1-\alpha)R_m I_t - \vartheta S_{t-1}\\
&\eta_t = \rho \eta_{t-1}+ (1-\rho)S_{t-1} \\
&\vartheta = b_0+\beta\eta_t \\
&S_t = \hat{f}_s(u_t,\vartheta),
\label{eq2:adaptsn}
\end{align}
where $\alpha,\gamma$ are parameters related to the temporal dynamics, $\alpha = \exp(-dt/\tau_m)$ and $\rho = \exp(-dt/\tau_{adp})$, $\vartheta$ is a dynamical threshold comprised of a fixed minimal threshold $b_0$ and an adaptive contribution $\beta \eta_t$;  
$\rho$ expresses the single-timestep decay of the threshold with time-constant $\tau_{adp}$. The parameter $\beta$ is a constant that controls the size of adaptation of the threshold; we set $\beta$ to 1.8 for adaptive neurons as default. Similarly, $\alpha$ expresses the single-timestep decay of the membrane potential with time-constant $\tau_m$.

The SRNNs were trained using BPTT, various spiking neuron models with plastic time-constants and with various surrogate gradients. Apart from the SSC and SHD datasets, analog input values are encoded into spikes either using spikes generated by a level-crossing scheme (ECG) or by directly injecting a proportional current into the first spiking layer (S-MNIST, PS-MNIST, SoLi, TIMIT, GSC). To decode the output of the network, we used one of two methods: either spike-counting over the whole time-window, for the (P)S-MNIST task, non-spiking LIF neurons (TIMIT, SHD, SoLi, and GSC), or spiking ALIF neurons (ECG). With spike-counting, classification is decoded from the sum of the output spikes as $\hat{y} = \text{softmax}(\sum_t S_{i,out}^t)$ where $S_{i,out}^t$ is the spike of the output neuron $i$ at time $t$. For either non-spiking LIF neurons and spiking ALIF neurons as outputs, a softmax classification is computed from the output neurons' membrane potential $u_{out,t}$ at each timestep as $\hat{y}_t = \text{softmax}(u_{out,t})$. For ECG, we used spiking ALIF neurons for outputs as they performed best, which we believe is related to the fact that this is the only task where classification switches within the sample - the spiking then functions effectively as resets.

We use a standard BPTT approach \cite{bellec2020solution} to minimize the cross-entropy (CE) or negative-log-likelihood (NLL) loss for each task using the Adam\cite{kingma2014adam} optimizer, where we unroll all input timesteps from end to the start. The error-gradient is calculated and accumulated through all timesteps after which the weights are updated. BPTT for the spiking neurons is calculated retrogradely along with the self-recurrence circuits. As shown in Figure \ref{fig:panel1}\textbf{d}, given an input sequence $X = {x_0,x_1,x_2, \dots,x_T}$, and a neuron with initial states $\{u_{h,0},u_{o,0},S_{h,0},S_{o,0}\}$, we obtain for each timestep $t\in\{0,T\}$ the spiking neuron states $\{u_{h,t}, S_{h,t}, u_{o,t}, S_{o,t} $\}, where $S_{h,t}$ refers to a neuron firing-or-not in a hidden layer and $S_{o,t}$ to an output neuron (if spiking), and $u_{h,t}$ and $u_{o,t}$ denote hidden and output neurons' membrane potentials. We then obtain a classification $\hat{y}(t)$ either for each timestep or for the whole sequence $\hat{y}$ and an associated loss. In classification tasks with $C$ classes, the prediction probability of class $c$ --  $\hat{y_c}$ is computed after having read the whole sequence, and  then the loss of the network is calculated as $\mathcal{L} = \sum_{c=1}^C y_c \log \hat{y_c}$, where $y_c$ is the target probability of class $c$. In streaming tasks (ECG, SoLi), the total loss is computed as the sum of the loss at each timestep -- $\mathcal{L} = \sum_{t=1}^T \mathcal{L}_t$. For the BPTT-derived gradient, we compute
$\frac{\partial \mathcal{L}}{\partial z} = \hat{y} - y$,
and for recurrent weights $W_{h2o}$, we compute $\frac{\partial \mathcal{L}}{\partial W_{h2o}} = \frac{\partial \mathcal{L}}{\partial z} \sum_{t'}^T \frac{\partial S_{o,t'}}{\partial W_{h2o}}$, where each term can be computed at each timestep $t'$ as $\frac{\partial S_{o,{t'}}}{\partial W_{h2o}} = \frac{\partial S_{o,{t'}}}{\partial u_{o,t'}} \frac{\partial u_{o,t'}}{\partial W_{h2o}}+\sum_{\xi=0}^{{t'}-1} \frac{\partial S_{o,{t'}}}{\partial u_{t'}} \frac{\partial u_{o,t'}}{\partial u_{o,\xi}} \frac{\partial u_{o,\xi}}{\partial W_{h2o}}$ and 
 
  $ \frac{\partial S_{o,{t'}}}{\partial W_{h2h}}= \sum_{\xi=0}^{{t'}} \frac{\partial S_{o,{t'}}}{\partial u_{h,\xi}}\frac{\partial u_{h,\xi}}{\partial W_{h2h}}$, and where $W_{h2h}$ refers to weights between neurons in the hidden layers, and $W_{h2o}$ to weight between hidden and output neurons.
The discontinuous spiking function enters the gradient as the term $\frac{\partial S}{\partial u}$, and here we use the differentiable surrogate gradients \cite{neftci2019surrogate}.  

For the Multi-Gaussian surrogate gradient, we found effective parameter values $h=0.15$ and $s=6.$ based on a grid search, and we set $\sigma$ to $0.5$. The standard surrogate gradients were defined following \cite{neftci2019surrogate}, with the Linear surrogate gradient as $\hat{f_s}'(u_{t}|\vartheta) = ReLU( 1-\alpha_{linear} |u_t-\vartheta|)$; the SLayer\cite{shrestha2018slayer}  gradient  as $\hat{f_s}'(u_{t}|\vartheta) = \exp( -\alpha_{slayer} |u_t-\vartheta|)$, and the Gaussian surrogate gradient as $\hat{f_s}'(u_{t}|\vartheta) = \mathcal{N}( u_t|\vartheta,\sigma_G)$; for all gradients, $\alpha$ is positive. We optimized all surrogate gradients hyperparameters in the experiments using grid searches; in the experiments we used  $\alpha_{linear} = 1.0$, $\alpha_{slayer} = 5.0$, and $\sigma_G = 0.5$. 

\paragraph{Network initialization.}  Compared to ANNs, SRNNs require initializing both weight and also the spiking neurons' hyperparameters (i.e, neuron type, time constants, thresholds, starting potential). We randomly initialize the time constants following the normal distribution $(\mu,\sigma)$ with per-layer specific parameters given in Table \ref{tab:params}. For all neurons, the starting value of the membrane potential is initialized with a random value distributed uniformly in the range[0,$\vartheta$]. The bias weights of the network are initialized as zero and all feedforward weights are initialized using Xavier-uniform initialization; weights for recurrent connections are initialized as orthogonal matrices. We compared networks with constant, uniform, and normal initializers for the time-constants and found that the normal initializer achieved the best performance (Figure \ref{fig:SoLi-ini}).

For the various tasks, the loss-function, sequence length, maximum number of epochs, learning rate and decay schedule, and minibatch-size are specified in Table \ref{tab:params}. Unless specified otherwise, the network architecture consists of inputs densely connected to one or more fully recurrently connected layers of spiking neurons connected to a layer of output neurons, as illustrated in Figure \ref{fig:panel1}b. For the {\bf ECG} task, the QTDB dataset \cite{laguna1997database} consists of two channels of ECG signals. We apply level-crossing encoding \cite{lichtsteiner2008128} on the normalized ECG signal to convert the original continuous values into a spike train, where each channel was transformed into two spike trains representing a value increasing or decreasing event respectively. The level crossing encoding we used is defined as 
\begin{align*}
&S_{+} = 
\begin{cases}
    1,& \text{if } x_t-x_{t-1}\geq L_{+}\\
    0,              & \text{otherwise}
\end{cases}, 
&S_{-} = 
\begin{cases}
    1,& \text{if } x_{t-1}-x_t-\geq L_{-}\\
    0,              & \text{otherwise}
\end{cases}
\end{align*}
where $S_{+},S_{-}$ denote spikes for the positive and negative spike-train respectively and we used $L_{+} = 0.3$ and $L_{-} = 0.3$. 

For the {\bf Spiking Heidelberg Dataset (SHD)}, the audio records were aligned to 1s by cutting or completing with zeros. As in \cite{cramer2019heidelberg}, two speakers were held out for the test dataset, and 5\% of samples from other speakers were also added into the test dataset. The training dataset thus comprised of 8156 samples and test dataset contains 2264 samples. For the {\bf Spiking Speech Command Dataset}, 
the speech commands were also uniformly aligned to 1s with a 250hz sampling frequency, and the dataset was randomly split into training, validation and test dataset with a ratio of 72\%-8\%-20\% respectively.
For the {\bf SoLi dataset}, the sequence of 40 Range-Doppler images (RDI) was fed into the model frame-by-frame as input and split into training and testset as in \cite{wang2016interacting}. 
The original RDIs have 4 channels, but we found empirically that using one channel was sufficient. For the SoLi task, the first layer of the SRNN, we use a feedforward spiking dense layer, followed by a recurrent layer. As in \cite{wang2016interacting}, separate networks were trained for per-frame accuracy (Acc$_s$) and per-sequence accuracy (Acc$_c$), for the streaming and classification version of the task respectively. In the {\bf S-MNIST} tasks, the network read the image pixel by pixel; for the {\bf PS-MNIST} task, pixels are read into the network using a sliding window of size 4 with stride 1. For both tasks, the pixel value is fed into the network directly as injected current into the neurons of the first hidden layer as a fully connected layer with its own weights.  We use the {\bf Google Speech Command}  v1 \cite{warden2018speech}. 
For preprocessing, Log Mel filters and their first and second-order derivatives are extracted from raw audio signals using Librosa \cite{mcfee2015librosa}. For the FFTs, a window of 30ms and a hop of 10ms is used. The timestep of the simulation is 10 ms. We calculate the logarithm of 40 Mel filters coefficient using the Mel scale between 20 Hz and 4kHz. Additionally, spectrograms are normalized to ensure that the signal in each frequency has a variance of 1 across time; we then selected the first three derivative orders as three distinct input channels. The input to the SRNN is thus a sequence of 101 frames, where each frame comprises of a 40-by-3 matrix. 
 

The {\bf TIMIT} database contains 3696 and 192 samples in training and test data respectively. We preprocessed the original audio data as in \cite{bellec2020solution} using MFCC encoding; 10\% of the training dataset was randomly selected as validation dataset, and the network was trained on the remainder. Similar to bi-directional LSTMs, we use a bi-directional Adaptive SRNN for this task, illustrated in Figure \ref{fig:compcost_SI}\textbf{a}: we use  two SRNN layers in the network, reading the sequence from the forward and backward direction respectively. The mean of these layer's output is then fed into the last layer, an integrator, to generate the class prediction.

\bibliography{sample}



\section*{Acknowledgements (not compulsory)}


BY is funded by the NWO-TTW Programme “Efficient Deep Learning” (EDL) P16-25.
The authors gratefully acknowledge the support from the organizers of the Capo Caccia Neuromorphic Cognition 2019 workshop and Neurotech CSA, as well as Jibin Wu and Saray Soldado Magraner, for helpful discussions.

\section*{Author contributions statement}
B.Y., F.C. and S.B. conceived the experiment(s),  B.Y. conducted the experiment(s), B.Y., F.C. and S.B. analysed the results. All authors reviewed the manuscript. 

\section*{Additional information}
 \textbf{Competing interests.} The authors declare no competing interests. 



\newpage


\renewcommand{\thepage}{S\arabic{page}} 
\renewcommand{\thesection}{S\arabic{section}}  
\renewcommand{\thetable}{S\arabic{table}}  
\renewcommand{\thefigure}{S\arabic{figure}}
\renewcommand{\figurename}{Figure}
\setcounter{figure}{0} 
\setcounter{table}{0} 
\section*{Supplementary Information}



\begin{figure*}[hbp]
    \centering
    \includegraphics[width=\textwidth]{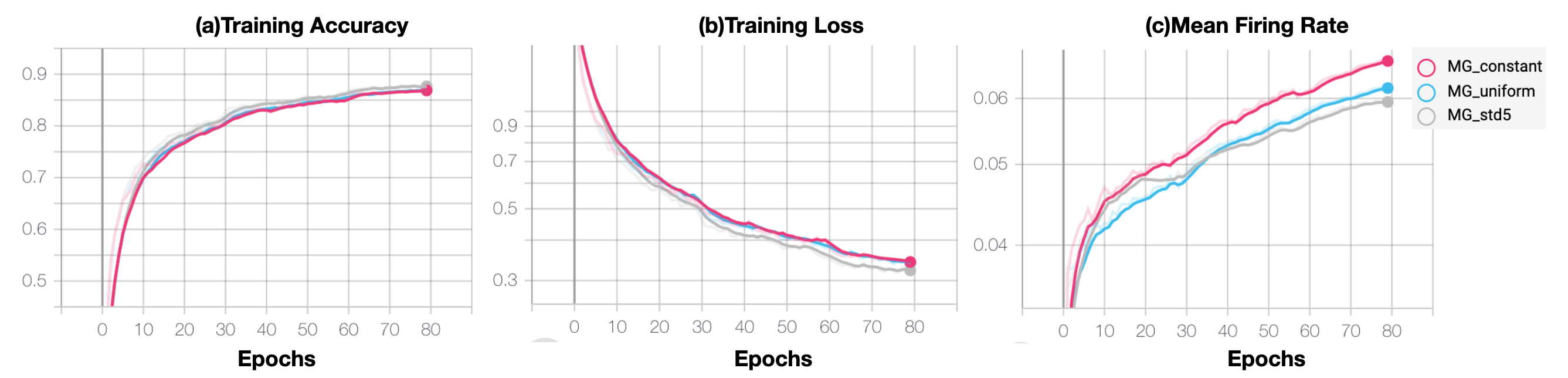}
    \caption{\textbf{Effects of different time constant initialization schemes on network training and performance on the SoLi dataset.} \textbf{a}, Training accuracy \textbf{b}, Training Loss \textbf{c}, Mean Firing rate of the network. The MG$_{constant}$ is the network where $\tau$ is initialized with a single value; for  MG$_{uniform}$ the network is initialized with uniformly distributed time-constants near the single value of MG$_{constant}$; for MG$_{std5}$, a normal distribution with std 5.0 is used near the same single value.}
    \label{fig:SoLi-ini}
\end{figure*}

\begin{table}[hb]
\centering
\caption{\textbf{Details of parameters used for initialization and optimization for each task.} Time constants refer to successive layers in the architecture; the layer output is last numbered layer -- the numbers between brackets denote the mean and std used for initializing the time constants with Gaussian distribution $N(\mu;\sigma)$ where the $\mu$ is the mean of the time constant and $\sigma$ is std. LIF non-spiking output layers use only the $\tau_m$ parameter. For the Loss, NLL denotes negative log-likehood and CE cross-entropy. The learning rate decay schedules denote the amount of decay (multiplicative factor) and the epoch when the decay is applied (in case of Step decays); for SMNIST and PSMNIST, a linear decay to zero is applied.}
\begin{tabular}{|l|l|l|l|l|l|l|l|l|l|}
\hline
\multicolumn{2}{|l|}{task}                 & ECG & SHD & SSC & GSC & SMNIST & PSMNIST & SoLi & TIMIT \\ \hline
\multicolumn{2}{|l|}{seq length}           &1301 &250  &250 &101 &784        &784         &40  &500       \\ \hline
\multirow{6}{*}{time constant} & $\tau_m1$  &(20;.5) &(20;5)&(20;5) &(20;5) &(20;5) &(20;5)  & (20;5)     &(20;5)       \\ \cline{2-10} 
                              &$\tau_{adp}1$&(7;.2)&(150;10)&(150;50) &(150;50) &(200;50)    &(200;50)  &(20;5)      &(200;5)       \\ \cline{2-10} 
                               & $\tau_m2$  &(20;.5)&(20;5)&(20;5)&(20;5) &(20;5)    &(20;5)    & (20;5)     &(20;5)       \\ \cline{2-10} 
                              &$\tau_{adp}2$&(100;1)&(150;10)&(150;50) &(150;50)&(200;50)&(200;50)  &(20;5)      &(200;50)       \\ \cline{2-10} 
                               &$\tau_m3$   & -  &(20;5)&(20;5)&(20;5)&(20;5)  &(20;5)    &(20;5)      &(3;1)  \\ \cline{2-10} 
                              &$\tau_{adp}3$& -  &  -  &  -   & -   &(200;50) &(200;50)   &   -   &  -     \\ \hline
\multicolumn{2}{|l|}{learning rate}        &1e-2 &1e-2 &1e-2 &1e-2&1e-2        &1e-2 &2e-2 &1e-2   \\ \hline
\multicolumn{2}{|l|}{Loss}                 &NLL &CE &CE &CE&CE    &CE     &NLL  &CE   \\ \hline
\multicolumn{2}{|l|}{minibatch size}        &64  &64   &32  &64&256        &256         &128 &16    \\ \hline
\multicolumn{2}{|l|}{epochs}               &400 &20  &150 &70 &300        &300         &150  &200       \\ \hline
\multicolumn{2}{|l|}{lr decay}             &.5per50    &.5per20  &.5per10 &.75per20 &        &         &.75per50  &.5per100       \\ \hline
\multicolumn{2}{|l|}{lr decay type}        &Step    &Step  &Step &Step &Linear        &Linear        &Step  &Step       \\ \hline
\end{tabular}
\label{tab:params}
\end{table}

\begin{figure}[ht!]
\centering

\includegraphics[width=\textwidth]{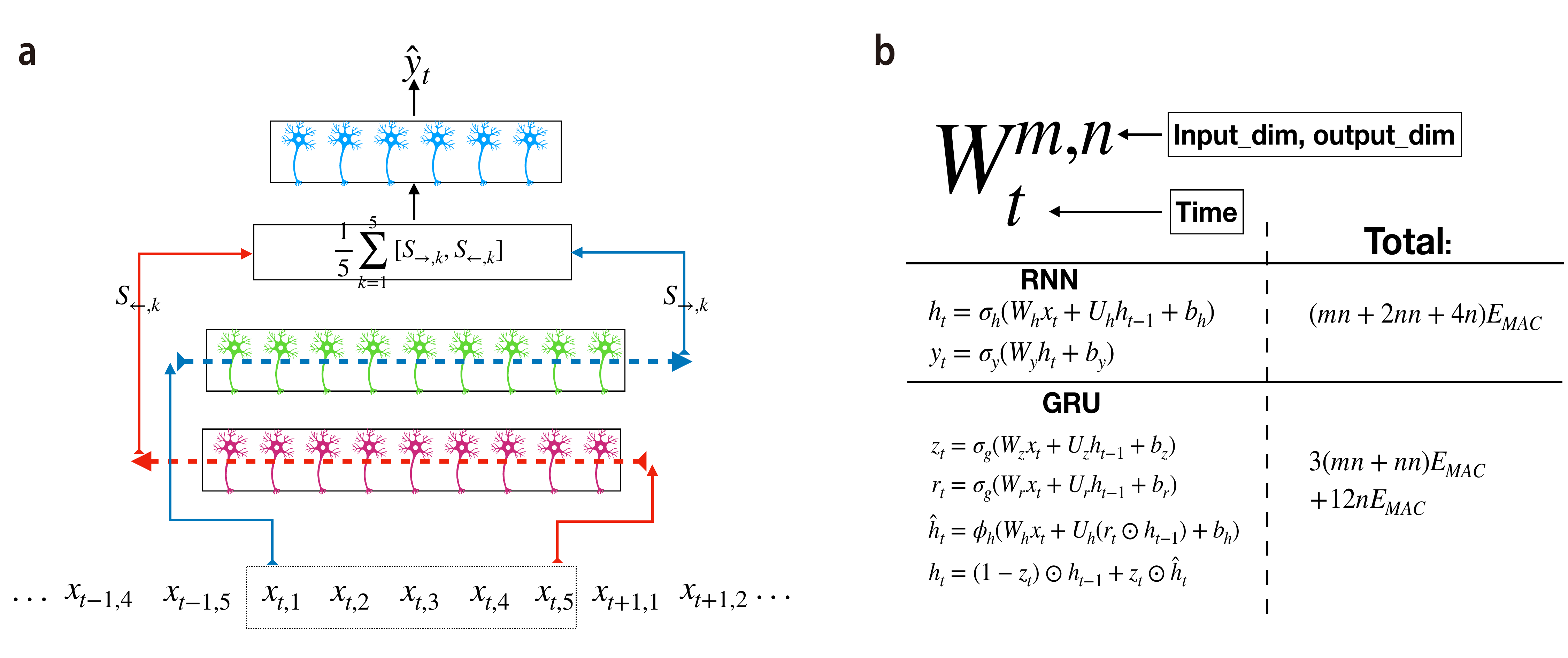}
\caption{\textbf{a}, Bi-directional SRNN architecture. \textbf{b}, Computational cost computation of different layers for regular RNNs and GRU units. The computational complexity calculation follows \cite{hunger2005floating}.} 
\label{fig:compcost_SI}
\end{figure}

\begin{table*}[t]
\small
  \centering
  \begin{tabu}{|l|l|r|r|r|r|r|}
  \hline
   Task                       & Method  & Accuracy &Energy$_{nn}$/s &Energy ratio & Error ratio & Efficiency\\ \hline
 \multirow{8}{*}{\bf ECG-qtdb}& Adaptive SRNN  &\bf \em 85.9\% &325.7           &1x           &1x     & 1x\\ \cline{2-7}
                              & LIF SRNN           &  75.5\%  &179.9            &.55 x       &1.67x & .91x\\ \tabucline[1pt]{2-7} 
                              & RELU SRNN          &  86.4\%  &5784.6         &17.8x        &.93x   &16.5x\\ \cline{2-7}   
                              & LSTM           &  78.9\%  &20422.8       &62.7x        &1.43x  &89.6x\\ \cline{2-7}
                              & GRU          &  77.3\%  &15400        &47.2x        &1.54x  &72.7x\\ \cline{2-7}
                              & Vanilla RNN  &  74.8\%  &9597.6          &29.5x        &1.71x  &50.5x\\ \cline{2-7}
                              & Bidirectional LSTM$_{290}$ &  80.76\%& 563580        &1729.9x            &1.31x  &2266.2x\\ \tabucline[1pt]{0-7} 

 \multirow{2}{*}{\bf SMNIST}  & Adaptive SRNN &\bf \em 98.7\%  &   8250.3         &1x           &1x     & 1x\\  \tabucline[1pt]{2-7} 
                              & RELU SRNN         &  98.99\%   &  487623.8       &59.1x        &.74x     &43.4x\\  \tabucline[1pt]{0-7} 
 \multirow{2}{*}{\bf PSMNIST}  & Adaptive SRNN &\bf \em 94.32\%  & 7775.1           &1x           &1x     & 1x\\  \tabucline[1pt]{2-7} 
                              & RELU SRNN      &  93.47\%   & 487623.8       &62.7x        &1.1x     &69.0x\\  \tabucline[1pt]{0-7} 

 \multirow{4}{*}{\bf SHD}& Adaptive SRNN &\bf \em 87.81\%  &3515.7           &1x           &1x     & 1x\\  \tabucline[1pt]{2-7} 
                              & RELU SRNN       &  88.93\%   &442097.2          &125.8x           &.91x      &114.5x\\ \tabucline[1pt]{2-7} 
                              & Bidirectional LSTM &  87.2\%   &3468215.52        &986.5x       &1.05x    &1035.8x\\ \tabucline[1pt]{2-7} 
                              & CNN\cite{cramer2019heidelberg} &  92.4\%   &  --       &--x        &--x     &--x\\ \tabucline[1pt]{0-7}

\multirow{2}{*}{\bf SSC}  & Adaptive SRNN &  74.18\%  &10154.1   &1x           &1x     & 1x\\  \tabucline[1pt]{2-7} 
                              & RELU SRNN &  74.36\%   &  2373918. &234.0x    &.99x     &231.7x\\  \cline{2-7} 
                              & LSTM\cite{cramer2019heidelberg} &  73.1\%   &  --       &--x        &--x     &--x\\  \cline{2-7} 
                              & CNN\cite{cramer2019heidelberg} &  77\%   &  --       &--x        &--x     &--x\\  \tabucline[1pt]{1-7} 

\multirow{2}{*}{\bf SoLi}  & Adaptive SRNN &  79.8\%  &13,804    &1x           &1x     & 1x\\  \tabucline[1pt]{2-7} 
                                      &Vanilla RNN &  63.6\%  &4101324.8 &297.1x       &1.80x  & 524.8x\\  \tabucline[1pt]{2-7} 
                                      &GRU &  79.20\%  &6580531.2 &476.7x       &1.03x  & 491.1x\\  \tabucline[1pt]{2-7}
                                      &LSTM&  79.99\%  &8360972.8  &605.7x       &0.99x  & 599.6x\\  \tabucline[1pt]{2-7} 
                                 &RELU SRNN&  79.8\%    &  3283950.9       &237.9x        &1.x     &237.9x\\  \tabucline[1pt]{0-7} 
                              
\multirow{2}{*}{\bf TIMIT}  & Adaptive SRNN & 66.13\%  & 10626.8           &1x           &1x     & 1x\\  \tabucline[1pt]{2-7} 
                              & RELU SRNN   &  --\%    &11861974          &175.2x        &--x     &--x\\  \tabucline[1pt]{0-7} 
\multirow{3}{*}{\bf GSC}  & Adaptive SRNN & 92.12\%  & 4120.3           &1x           &1x     & 1x\\  \tabucline[1pt]{2-7} 
                              & RELU SRNN   &  --\%    &11861974          &167.5x        &--x     &--x\\  \tabucline[1pt]{2-7} 
                              & CNN\cite{wong2020tinyspeech}   &  92.4\%    &80600         &19.6x        &0.96x     &18.8x\\  \tabucline[1pt]{0-7} 
\end{tabu}
     
  \caption{\textbf{Performance and relative energy consumption for various models.} For comparison, the average energy consumption of each input timestep and error rate of Adaptive SRNN was set to unit value. The energy/error ratio is defined as the ratio of the energy/error compared to the adaptive SRNN. We define the efficiency as the product of energy and error ratio. In the ECG task, the Bidirectional LSTM$_{290}$ is a network with 2 bidirectional LSTM with 120 and 40 hidden neurons respectively followed by three dense layer with 100, 20 and 6 hidden neurons. For the ECG tasks, the comparison networks (LSTM, GRU, vanilla RNN and LIF SRNN) share the same architecture with 36 hidden neurons each. In the SHD task, the bidirectional LSTM is a 2-layer bidirectional network with 128 neurons each, followed by a dense layer with 100 neurons. In the SHD and SSC tasks, the temporal bin-width for the LSTM and CNN networks  \cite{cramer2019heidelberg} is set to 10ms with a sampling frequency of 100Hz.  For SoLi, all networks (vanilla RNN, GRU, and LSTM) use the same network structure as the SRNN, with a recurrent layer with 512 neurons followed by a dense layer with 512 neurons. All CNNs in the table read the whole sequence at once, and their computation cost is computed as the average cost per timestep. Dashes denote where values are not available either because of lacking network architecture details or lack of convergence (ReLu SRNN for TIMIT and GSC).}
  \label{tab:sota-ratio}
\end{table*} 
\end{document}